\documentclass[letterpaper, 10 pt, conference]{ieeeconf}  %

\IEEEoverridecommandlockouts                              %

\usepackage{mathptmx} %
\usepackage{times} %
\usepackage{amsmath} %
\usepackage{amssymb}  %
\usepackage{graphicx}
\usepackage{subfigure}

\usepackage{float}

\usepackage{multirow}
\usepackage{diagbox}
\usepackage{booktabs}

\usepackage{balance}

\makeatletter
\def\blfootnote{\xdef\@thefnmark{}\@footnotetext}
\makeatother

\usepackage{tikz}
\usetikzlibrary{matrix,calc}

\title{\LARGE \bf
Interactive Expressive Motion Generation\\Using Dynamic Movement Primitives
}

\author{Till Hielscher, Andreas Bulling, Kai O. Arras%
\thanks{T. Hielscher and K.O. Arras are with the Socially Intelligent Robotics Lab, Institute for Artificial Intelligence, University of Stuttgart, Germany {\tt\small till.hielscher@ki.uni-stuttgart.de}}%
\thanks{A. Bulling is with the Collaborative Artificial Intelligence Group, Institute for Visualization and Interactive Systems, University of Stuttgart, Germany}%
\thanks{The authors thank the International Max Planck Research School for Intelligent Systems (IMPRS-IS) for supporting Till Hielscher.}
}

\newcommand{\acceptance}{%
\begin{tikzpicture}[overlay, remember picture]
\path (current page.north) ++(0,-1cm) node[align=center] {
This paper has been accepted for publication at the 2025 IEEE/RSJ International Conference on Intelligent Robots and Systems (IROS).
};
\end{tikzpicture}
}
\newcommand{\isArxiv}[2]{#1} %

\begin{document}

\maketitle
\isArxiv{\acceptance}{}
\thispagestyle{empty}
\pagestyle{empty}

\begin{abstract}

Our goal is to enable social robots to interact autonomously with humans in a realistic, engaging, and expressive manner. The 12 Principles of Animation are a well-established framework animators use to create movements that make characters appear convincing, dynamic, and emotionally expressive. This paper proposes a novel approach that leverages Dynamic Movement Primitives (DMPs) to implement key animation principles, providing a learnable, explainable, modulable, online adaptable and composable model for automatic expressive motion generation. DMPs, originally developed for general imitation learning in robotics and grounded in a spring-damper system design, offer mathematical properties that make them particularly suitable for this task. Specifically, they enable modulation of the intensities of individual principles and facilitate the decomposition of complex, expressive motion sequences into learnable and parametrizable primitives.
We present the mathematical formulation of the parameterized animation principles and demonstrate the effectiveness of our framework through experiments and application on three robotic platforms with different kinematic configurations, in simulation, on actual robots and in a user study.
Our results show that the approach allows for creating diverse and nuanced expressions using a single base model.

\end{abstract}

\section{Introduction}
    As robots are increasingly deployed in places where they interact directly with humans, there is a growing need for them to possess social interaction capabilities that are perceived as pleasant and engaging. A key aspect of this trend is non-verbal interaction and intent communication, using, for example, motion through space or expressive body language \cite{breazeal_social_2004,kato_may_2015,embgen_robot-specific_2012}.
    Gestures are important in conveying intent, both as a complement to speech and as an independent mode of communication through deictic or symbolic movements \cite{bartneck_human-robot_2020}. Related work indicates that variations in the execution of expressive motion can significantly influence perception. For instance, in \cite{bremner_conversational_2009}, the same gesture was found to be interpreted differently as a function of timing and smoothness. Additionally, adapting a robot's posture has been shown to enhance its expressiveness \cite{beck_interpretation_2010}, while modulating the velocity and extent of movements can improve the perception of lifelikeness and emotional expression \cite{bartneck_human-robot_2020}.
    These findings confirm the established practice of animators who modulate expressive motion to enhance characters' perceived realism and engagement. To achieve this, they commonly employ the 12 Principles of Animation \cite{thomas_illusion_1995}, a foundational framework for bringing animated characters to life.
    
    In this paper, we leverage techniques from robot imitation learning to implement these principles, aiming to achieve automatic expressive robot motion generation through a learnable, explainable, modulable, online adaptable, and composable model.
    Concretely, we make the following contributions:
      \begin{enumerate}
        \item We introduce Dynamic Movement Primitives (DMPs) as a new unifying implementation of the Principles of Animation, and show that the model of motor primitives as stable dynamical systems has favorable properties for the task of expressive motion generation.
        \item Using DMPs, we encode a relevant selection of principles that can be individually modulated to shape a learned reference motion.
        The experiments in simulation, on three different robots, and a user study demonstrate that the approach allows for effectively creating diverse expressive motion through intuitive modulation using a single base model.
    \end{enumerate}
    
    \begin{figure}
    \vspace{-6mm}
        \centering
        \includegraphics[width=0.95\linewidth]{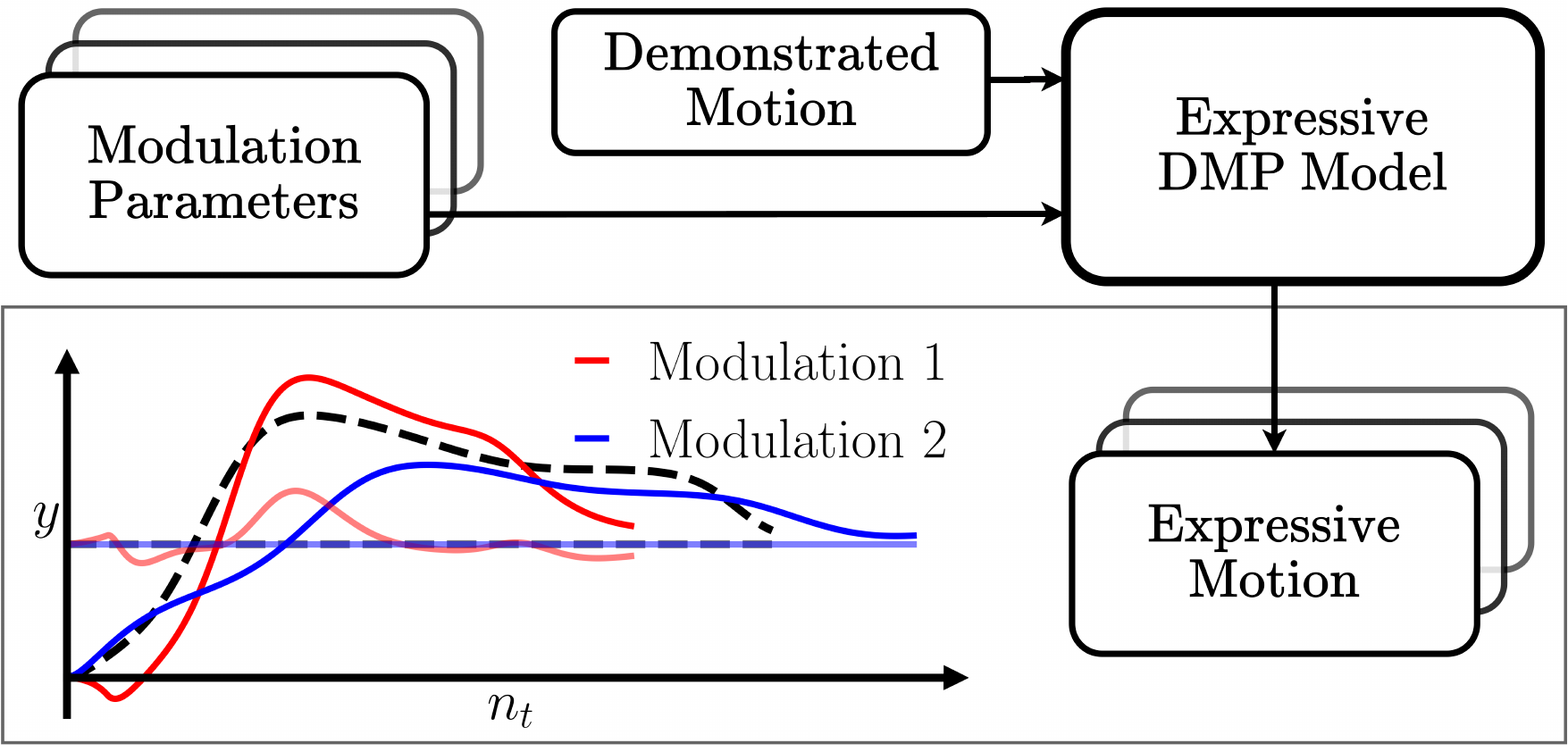}
        \vspace{-8mm}
        \caption{Our framework takes demonstrated trajectories and modulation parameters as input and generates expressive robot motion. Shown are two joint angles over time with black (dashed) demonstrations and two example modulations in red and blue. Some animation principles define primary and secondary motion, which are shown as opaque for the primary and translucent for the secondary joints.}
        \label{fig:framework_overview}
        \vspace{-3mm}
    \end{figure}

\section{Related Work}
    This section discusses related work in expressive motion representation, generation, and DMPs.

    \subsection{Expressive Motion Representation and Generation}

        Prominent approaches for the representation of expressive human motion are the 12 Principles of Animation \cite{thomas_illusion_1995}, discussed in Sec. \ref{sec:method}.A, the Laban dance notation \cite{guest_labanotation_2005, laumond_dance_2016, abe_use_2017}, an accurate expressivity description scheme used in choreography planning, and motion descriptors \cite{larboulette_review_2015}, methods that compute low-level kinematic and geometric descriptors or high-level body, space, shape, and effort descriptors. Examples include acceleration, convex hulls, rotation, action presence, covered distance, extensiveness, or flow effort.

        For the generation of expressive robot motion, methods range from low-level control to learning techniques using generative AI. See also the survey in \cite{venture_robot_2019}. The Principles of Animation have been used in several works:
        In \cite{takayama_expressing_2011}, the researchers selected two animation principles for a robot to show forethought before actions and express a reaction to a task's outcome.
        In \cite{van_breemen_bringing_2004}, the authors designed animations based on principles for the \textit{iCat} robot. In subsequent work, they used these animations to compose sequences from the individual animations \cite{van_breemen_animation_2004}.
        The results in \cite{zhou_expressive_2017} show the importance of the timing animation principle in the robot intent expression.
        The authors of \cite{ribeiro_illusion_2012} provide a general overview of how animation principles relate to robotics. They apply a selection of principles to emphasize emotion-related gestures for the \textit{EMYS} robot and verify the expressivity of the outcome with a user study. They extend their work with exemplary robot animations in \cite{ribeiro_practice_2020}. In \cite{szafir2014communication}, the authors propose a method for intent communication of a UAV by modulating flight trajectories based on three principles, each implemented with an individual mathematical model to adjust relevant motion attributes. They show a positive impact on users with respect to usability, safety, and naturalness.
        
        An inverse dynamics control approach was proposed in \cite{ramos_dancing_2015} to efficiently compute control commands for the purpose of expressive dancing of a humanoid robot.
        In \cite{toyoguchi_human-centered_2023}, an optimization-based controller is employed whose cost function is adapted by perceptual cues (personality and mood) of the human to guide the system toward desired expressivity. 
        By applying null-space decompositions, the authors of \cite{claret_exploiting_2017} and \cite{hagane_robotic_2022} exploit the redundancies of the robot kinematic structure to make the generated motion expressive.
        
        Multiple previous works use learning from demonstration (LfD) to align robotic actions with expressive actions from human experts. Some researchers learn cost functions \cite{huang_conditional_2023} or high-level plans \cite{mahadevan_generative_2024}.
        Relevant to our work are policy learning approaches that map information spaces like perceptual data, system states, contextual cues, or domain-specific heuristics to action spaces. %
        In \cite{osorio_control_2022}, models are trained to map human motion features onto spatial velocities for the robot, which can then be controlled with a proportional controller.        
        Using reinforcement learning, whole-body control policies can be made expressive by setting an expressive goal during training \cite{cheng_expressive_2024}.
        Generally, as stated by \cite{saran_enhancing_2019}, enhancements to LfD can be made by including human social cues.        
        
        Generative artificial intelligence methods have recently been used for expressive robot motion generation. In \cite{osorio_generative_2024}, robot states are obtained from an expressive generator model. The model is trained by aligning the training to human expressive motion contrasted with neutral motion.
        In \cite{marmpena_data-driven_2022}, motion is mapped into an emotion representation space, the valence-arousal-dominance (VAD) space, and the generation of expressive motion is realized for the kinematic structure of a single robot by utilizing a conditional variational autoencoder.
        Using the same latent VAD representation, the authors of \cite{sripathy_teaching_2022} generate expressive motion for arbitrary kinematic structures by implementing a style discriminator function to evaluate trajectories trained with human feedback.

    \subsection{Dynamic Movement Primitives}

    Dynamic Movement Primitives are a group of methods for encoding, learning, and generating complex and adaptable motion in robotics. DMPs are based on the idea of encoding a point-to-point motion into a stable nonlinear dynamical system whose parameters are learned from demonstrated trajectories using, e.g., locally weighted regression or reinforcement learning \cite{ijspeert_movement_2002, ijspeert_learning_2002, schaal_learning_2005, ishikawa_policy_2008}. An important feature of DMPs is their sequencing ability, which allows for a macroscopic policy representation in which complex motor behaviors are broken down into simpler and reusable movement primitives.

    Various extensions of the original approach have been developed (see the survey in \cite{saveriano_dynamic_2023}) and include their ability to represent Cartesian orientations in higher dimensions with coupled degrees of freedom \cite{abu2015adaptation}, their ability to incorporate via-points \cite{zhoulearningIROS19} or task parameters \cite{ude2010task}, as well as their online adaptation behavior for tasks such as obstacle avoidance by introducing an additional coupling term \cite{hoffmann_biologically-inspired_2009, gams_coupling_2014}. Several works address the issue of smoothly joining a sequence of DMPs \cite{stulp_hierarchical_2011,kulvicius_joining_2012,daab_incremental_2024}.
    Further extensions are probabilistic movement primitives (ProMPs) using Gaussian-based hierarchical Bayesian models \cite{paraschos_probabilistic_2013} and probabilistic dynamic movement primitives (ProDMPs) \cite{li_prodmp_2023}, which combine ProMPs and DMPs to produce bound and smooth trajectory distributions capturing higher-order statistics.
        
    DMPs have been used in many applications, most prominently for motor skill learning in humanoids \cite{ude2010task,zhou_learning_2016,daab_incremental_2024} and robot manipulation \cite{abu2015adaptation}, e.g. for assembly \cite{rozo2024ebike}, but also for movement representation of self-driving cars \cite{huang_conditional_2023}, aerial \cite{tomic2014learning} or underwater field robots \cite{carrera2015cognitive}.
    Applications of DMPs in human-robot interaction focus mainly on physical HRI scenarios. Iori et al. \cite{iori_dmp-based_2023} introduce a coupling term to time goal reaching in handover scenarios. The coupling term uses the perceived distance from the handover point to influence the robot's movements to that point. By joining a language-conditioned semantic model with a control model based on DMPs, the authors in \cite{stepputtis_language-conditioned_2020} allow for abstractions in a manipulation task with language. In \cite{scheikl_movement_2024}, a diffusion model determines the parameters of a ProDMP that the authors use for trajectory generation that meets requirements from robot-assisted surgery regarding versatility and quality.

    In contrast to the works discussed above, our approach differs as follows:
    Our implementation of the animation principles is, to the best of our knowledge, the first one based on a unified, physically plausible mathematical model that automatically modulates individual principles. This approach overcomes the limitations of handcrafted modulations \cite{takayama_expressing_2011, van_breemen_bringing_2004, van_breemen_animation_2004, zhou_expressive_2017, ribeiro_illusion_2012, ribeiro_practice_2020} and ad-hoc models tailored to specific principles \cite{szafir2014communication}. Moreover, and in the spirit of LfD, our approach leverages imitation learning as a way to acquire reference motion.  
    Unlike recent deep learning methods, our approach is data-efficient, explainable, and naturally suited for complex and online adaptive motor behavior.
    With respect to the DMP literature, our work is the first to adopt DMPs for social HRI and to the task of expressive motion generation.

\section{Method}
\label{sec:method}
    We first briefly summarize the original DMP approach in \cite{ijspeert_movement_2002,ijspeert_dynamical_2013}, put it in context, and describe our extensions. Refer to \cite{saveriano_dynamic_2023} for a comprehensive treatment.
    
    For a single dimension, a discrete DMP is defined by a second-order nonlinear differential equation
    \begin{equation}
        \tau^2 \ddot{y} = \alpha \left( \beta \left( g - y \right) - \tau \dot{y} \right) + f(x).
        \label{eq:dmp}
  \end{equation}
    Here, $y$ is the position and $\dot{y}$, $\ddot{y}$ its derivatives, velocity and acceleration. The execution time is denoted by $\tau$, with $\delta t$ being the discretization between steps $n_t = 1, \dots N_t$ with $N_t = \frac{\tau}{\delta t}$. The goal position of the system is given as $g$. The system gain parameters $\alpha = 4 \beta$ are chosen to ensure convergence to $y = g$, $\dot{y} = 0$. Dependent on the phase variable $x$, which resembles the time progression of the system, $f(x)$ is the system's forcing term made from a combination of $N_w$ Gaussian basis functions $\Psi$, equally distributed in time, with learnable weights $w$ as shown in Eq.~\ref{eq:forcing_term},

    \begin{equation}
        f(x) = \dfrac{\sum_{{i}=1}^{N_w} w_{i} \Psi_{i}(x)}{\sum_{{i}=1}^{N_w} \Psi_{i}(x)} \: x.
        \label{eq:forcing_term}
    \end{equation}

    Multi-dimensional problems with $n_{dim} = 1, \dots, N_{dim}$ can be addressed by assigning each dimension its own attractor system and forcing term. In contrast, all systems share the same phase variable $x$. Following this approach, the phase couples all dimensions.
    Each dimension of the multi-dimensional system has to be provided with a demonstration trajectory for weight computation and an individual goal.

    Learning DMPs in Eq. (\ref{eq:dmp}) amounts to estimating the weights using Locally Weighted Regression (LWR) \cite{schaal_constructive_1998}. Demonstrated trajectories are learned in a \emph{single} shot. Data can come from a 3D body pose tracker, motion profiles of an animator, or a planning algorithm.

    Besides the advantageous mathematical grounding as second-order ODEs, commonly used for modeling biomechanical systems in general and physics-based character animation, in particular, \cite{armstrong_dynamics_1985, wilhelms_using_1985}, DMPs come with a characteristic that is particularly useful for interactive, expressive motion generation: The inherent goal orientation %
    which ensures that gestures are naturally executed towards the person or object of interest in the scene. Given a visual detector for people or objects that provides $g$, the system interactively targets its motion towards that person or object, allowing online motion adaptions in real-time while maintaining the characteristic shape of the gesture. We denote this goal adaptability by $\mathcal{P}_\text{Goal}$.

    \subsection{Encoding Animation Principles}\label{sec:method_ani_principles}
        Initially developed for early animation and hand-drawn characters, the 12 Principles of Animation \cite{thomas_illusion_1995} remain a foundational framework for characters in 3D \cite{lasseter_principles_1998} and social robots. 
        Before introducing the parametrized principles using DMPs, we discuss the ones that are not relevant for expressive motion generation with robots as they do not apply to real rigid-body robot systems, are focused on the context, the animation technique or character design choices before animation:

        \emph{Appeal} addresses the character's appearance and charisma. It should ensure that the characters are engaging and interesting. 
        \emph{Solid Drawing} refers to an animator's drawing skills and is crucial for making characters appear three-dimensional and believable. %
        \emph{Squash and Stretch} deforms objects or body parts to give them a sense of flexibility and weight, like a softball that stretches and compresses when bouncing off. \emph{Staging} makes sure that ideas (e.g. actions, expressions) are clear by directing the audience's attention and properly presenting characters and objects in the scene. \emph{Straight Ahead Action and Pose to Pose} refers to two different animation techniques, either frame-by-frame or using frame interpolation between key poses.
          
        We now describe the remaining principles and how their individual modulation is realized using DMPs.
        Note that all modulated quantities replace their counterparts in the previously described formulation.
        \subsubsection{Arc $\mathcal{P}_{Arc}$}\mbox{}\\        
            Arcing influences how smooth and natural a movement appears. The number of DMP weights $N_w$ can adjust this motion characteristic. A lower $N_w$ results in less precise tracking of the demonstrated reference trajectory, leading to a broader arc. However, since $N_w$ is a parameter set during training, it is unsuited for online modulation. Instead, low-pass filtering of the weights can achieve a similar effect, attenuating their relative distribution. This is done by convolving a 1D Gaussian filter $G(\sigma)$ with the weights in each dimension. The modulation parameter $p_{\text{Arc}}>0$ controls the standard deviation $\sigma$ of the Gaussian kernel, effectively regulating the degree of smoothing.
            To obtain more sharply arced trajectories, we apply unsharp masking, an edge-enhancing technique from image processing that extracts high-frequency components by comparing the original and smoothed signals.
            Again, the modulation parameter $p_{\text{Arc}}<0$ controls the standard deviation $\sigma$ of the kernel.
            \begin{equation}
                {w}^{\,\prime} =
                \begin{cases}
                  \; G(p_{\text{Arc}}) * w        & \; p_{\text{Arc}} > 0 \\
                  \; w+(w-G(-p_{\text{Arc}}) * w) & \; p_{\text{Arc}} < 0
                \end{cases}
            \end{equation}
    
        \subsubsection{Anticipation $\mathcal{P}_{Ant}$}\label{sec:anticipation}\mbox{}\\        
            Anticipation prepares the audience for an action that is about to occur and creates more realistic and believable movements.
            Anticipation can be achieved by inverting the generated acceleration of the system for a short time window at the beginning of the execution. The time window can be adapted with the variable $T_{\text{Ant}}$, and the intensity of the anticipation motion is varied with the scaling factor $p_{\text{Ant}}$. With values $p_{\text{Ant}} > 0$, the anticipation motion is executed, where higher values lead to stronger anticipation motion.
            \begin{equation}
                {\ddot{y}}^{\,\prime} = - p_{\text{Ant}} \; \ddot{y} 
            \end{equation}    
            As anticipation should prepare for important parts of the motion, the application should only consider important dimensions or degrees of freedom. Therefore, inverting accelerations only apply to the $N_{\text{Ant}}$ most important dimensions. We determine importance as the differences between minimum and maximum $y$-values, where the dimension with the highest difference is chosen as the most important.
            Alternatively, the importance inference can be solved differently based on the user's given requirements or can be set manually.
        \subsubsection{Slow In and Slow Out $\mathcal{P}_{\text{Slow}}$}\mbox{}\\        
            This principle refers to how objects accelerate and decelerate, making their movements appear more natural. Animations are smoother when they start slowly, build speed, and slow down again.

            This can be achieved by modulating the phase, where
            slow phase decay leads to slower progression of the motion.
            Therefore, slow motion at the beginning and the end of the trajectory can be realized with slow phase progression at the beginning and the end.
            Unlike the original DMP theory, where the canonical system generates the phase $ \tau \dot{x} = -\alpha_x x$ \cite{ijspeert_dynamical_2013}, we use a linear phase during learning. This causes the basis functions to be equally distributed over time, leading to minimal average deviations favorable for later modulations introduced with our method. Possible tracking errors are negligible if $N_w$ is chosen to be high.

\begin{figure}[t]
                \vspace{4mm}
                \centering
                \subfigure[$\phi_{\text{Slow}}(n_t)$]{
                    \centering
                    \includegraphics[width=0.46\linewidth, trim={ 0 0 0 13pt},clip]{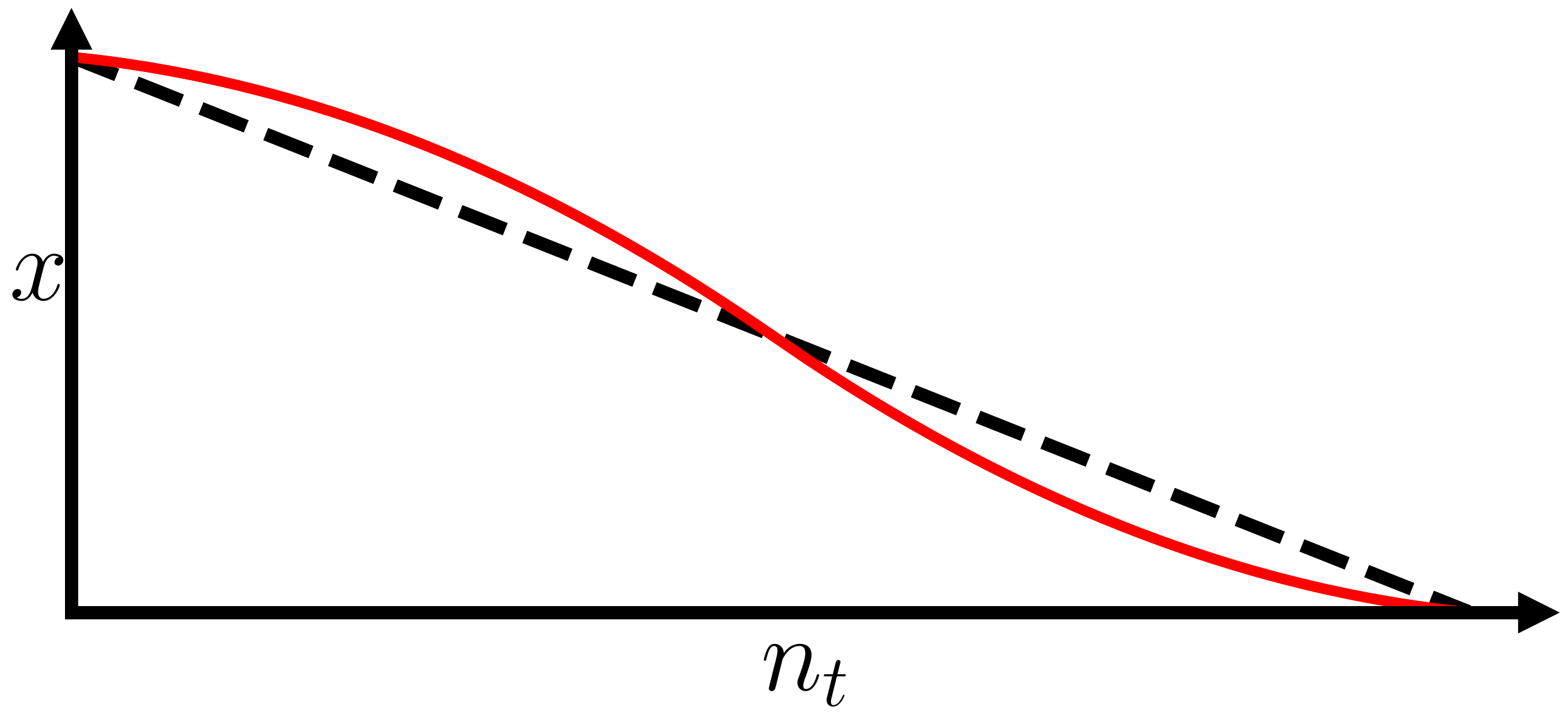}
                    \label{fig:phase_siso}
                }
                \subfigure[$\phi_{\text{Time}}(n_t)$]{
                    \centering
                    \includegraphics[width=0.46\linewidth, trim={ 0 0 0 13pt},clip]{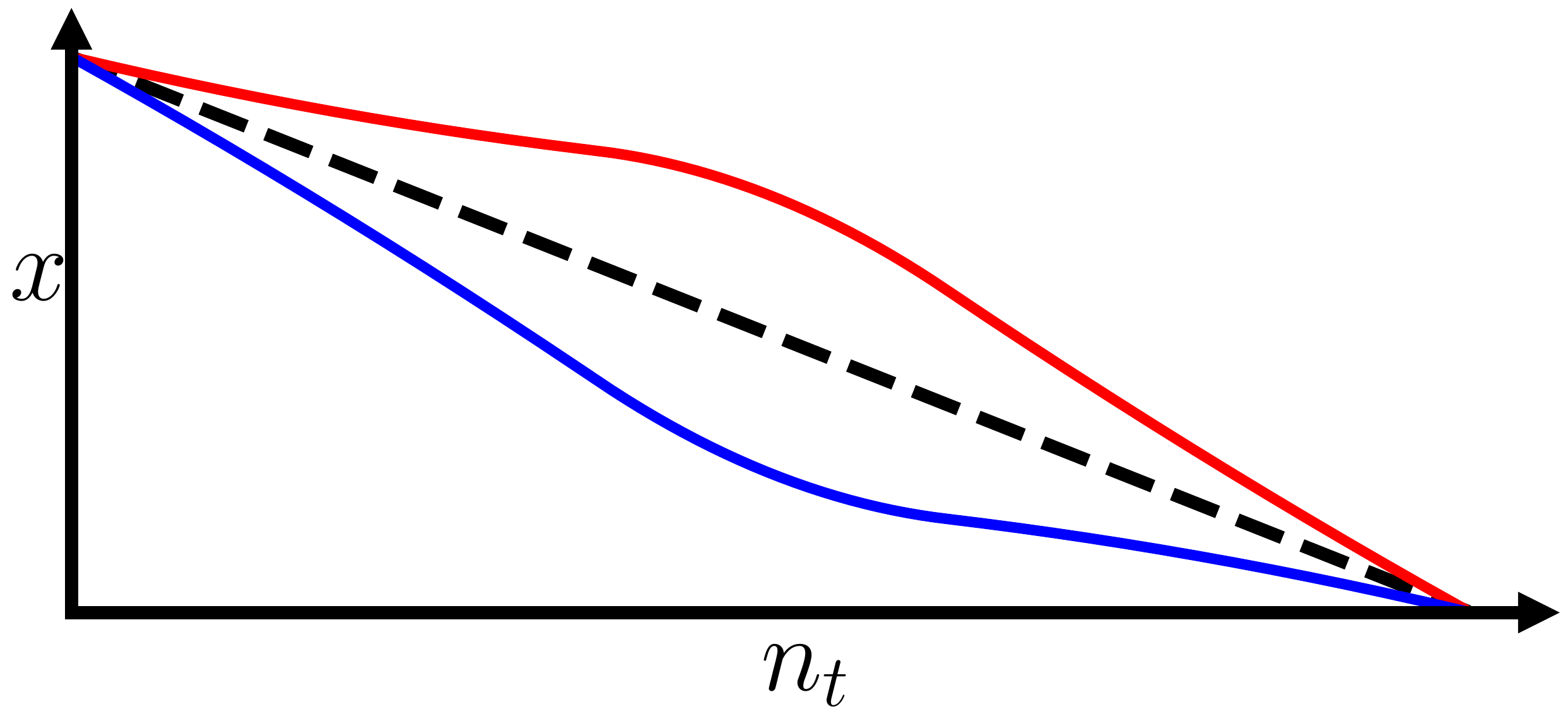}
                    \label{fig:phases_speed}
                }
                \caption{Phase adaptions for progression modulation with $\mathcal{P}_{Slow}$ and $\mathcal{P}_{Time}$. The linear phase, used during learning, is shown as a dashed line.}
                \label{fig:timing_phases}
                \vspace{-4mm}
            \end{figure}

            Ultimately, phase modulation is achieved by obtaining the phase from a modulation function $\phi_{\text{Slow}}(n_t)$. For the case of $\mathcal{P}_{\text{Slow}}$, an inverted sigmoid function can be used.
            The linear phase used during learning (dashed line) and the phase $\phi_{\text{Slow}}(n_t)$ are visualized in Fig.~\ref{fig:timing_phases}.a.
        \subsubsection{Timing $\mathcal{P}_{Time}$}\label{sec:timing}\mbox{}\\       
            Timing closely relates to the execution speed of a motion. Correct timing and, therefore, correct speed ensures that an action appears as it should in terms of character intent.    
            By scaling the system's execution time $\tau$, the time taken by the system to execute the motion can be extended or shortened. %
            Scaling $\tau$ directly affects the phase variable. As the system depends on the phase, an extended or shortened execution time resembles a modulation along the time axis. The scaling factor is denoted by $p_{\text{Time}}$.
            \begin{equation}
                \tau^{\,\prime} = p_{\text{Time}} \, \tau
            \end{equation}
            
            In addition to changing the system's execution time, another important aspect of the principle is the execution speed along the motion. Some motions should start slow and end fast, or vice versa.
            Making use of the phase modulation that is introduced for $\mathcal{P}_{\text{Slow}}$ enables progression modulation, which leads to modulated execution speeds along the trajectory.
            The phase modulation function $\phi_{\text{Time}}(n_t)$ produces a desired phase.
            Generally, this function can be designed arbitrarily based on given requirements to enable arbitrary progression modulation. For this work, we used a sector-based approach with spline interpolation.
            This leads to phases as visualized in Fig.~\ref{fig:timing_phases}.b, where the linear phase (dashed) is used during learning, and the two other phases resemble slow$\rightarrow$fast (red) and fast$\rightarrow$slow (blue) progression.
            
        \subsubsection{Exaggeration $\mathcal{P}_{\text{Exa}}$}\mbox{}\\        
            Exaggeration involves taking an action or expression to an extreme to increase the clarity or impact. It helps to convey the emotion or action more vividly to the audience.
            With DMPs, the forcing term addresses the action of the motion along the trajectory to the goal state. Modulating this action to be an overstatement or an understatement can be achieved by adjusting the influence of the forcing term within the attractor equation.
            Amplification of the induced action forces results in an overstatement and a reduction in an understatement.
            Therefore, scaling the forcing term with $p_{\text{Exa}}$ is appropriate to convey exaggeration. Values $p_{\text{Exa}} < 1$ lead to understatement and values $p_{\text{Exa}} > 1$ lead to overstatement.
            \begin{equation}
                f^{\,\prime} = p_{\text{Exa}} \; f
            \end{equation}
        \subsubsection{Secondary Action $\mathcal{P}_{\text{Sec}}$}\label{sec:secondary_action} \mbox{}\\
            Secondary actions support the primary action to add more life to the scene. For instance, a character walking might also swing their arms.

            This is realized by inducing velocities of the primary action into the position of another dimension.
            By doing this, fast movements in the primary action are especially respected and used for modulation.
            \begin{equation}
                y_{\text{target}}^{\,\prime} = y_\text{target} + \delta \, p_{\text{Sec}} \; \dot{y}_{\text{source}} 
            \end{equation}
            The primary action is linked to the \textit{source}, and the \textit{target} is the modulated dimension. Inverse relations can be specified with $\delta$, either $1$ or $-1$.
            For values $p_{\text{Sec}} > 0$, the secondary action is executed, with higher values leading to stronger secondary actions.
            Secondary actions are based on specific design choices and require knowledge of the robot's kinematic structure. For modulation, we assume suitable joint relationships given that specify \textit{source}, \textit{target}, and inverse relations.
            
        \subsubsection{Follow Through $\mathcal{P}_{\text{Follow}}$}\mbox{}\\        
            Parts connected to the main body drag and follow through with the motion of the main body after the main body stops, essentially creating overlapping actions in the grand scheme. 
            For this, the acceleration of the connected part is modulated by a scaled inverse acceleration of the main body.
            \begin{equation}
                \ddot{y}_{\text{target}}^{\,\prime} = \ddot{y}_{\text{target}} - \delta \, p_{\text{Follow}} \; \ddot{y}_{\text{source}}
            \end{equation}
            Here \textit{target} refers to the connected body's dimension and \textit{source} is the dimension corresponding to the main body. Inverse relationships can be addressed with $\delta$, either $1$ or $-1$.
            With values $p_{\text{Follow}} > 0$, the follow-through action is executed, where higher values lead to more pronounced movements.
            Note that the selection of \textit{target} and \textit{source} is dependent on the system's kinematic structure and possible intermediate degrees of freedom (DoFs). 
            Therefore, to apply modulation, the \textit{target} and \textit{source} must be in the same kinematic chain, with the \textit{source} higher in the hierarchy. The orientation of their rotation axes should be nearly identical within a defined threshold. Without a threshold, only perfectly aligned joints are modulated. Intermediate DoFs may need to be adjusted to meet this orientation threshold with specified limits.
              
        \subsubsection{Randomization $\mathcal{P}_{\text{Rand}}$} \mbox{}\\
            Movement randomization is not a core animation principle but enhances certain principles, making repeated motions appear more lifelike and natural. In our DMP approach, this can be easily achieved by varying the weights $w_i$ of the nonlinear term to generate colored motion noise around the learned reference trajectory.
            We add normally distributed random values $\epsilon$ to the DMP weights. As the magnitudes of the weights can vary strongly with the specific reference trajectory, the random values are scaled by the mean of the absolute weight values. The intensity of the movement randomization is then controlled by the parameter $p_{\text{Rand}}>0$.
            \begin{equation}
                w^{\,\prime} = w +  \Big(1+\tfrac{1}{N_w} \textstyle \sum_i |w_{i}| \Big)  p_{\text{Rand}} \; \epsilon, \quad \epsilon \sim \mathcal{N}(0, 1)
            \end{equation}
    
            TABLE~\ref{tab:parameters} summarizes the considered modulations and their parameters. They span a low-dimensional and meaningful space in which diverse and nuanced robotic expressions based on a reference motion can be created intuitively along the established dimensions of the Principles of Animation.
            The design process then amounts to adjusting the intensities of the parameters given the respective ranges.

            \begin{table}[t]
            \vspace{3mm}
            \centering
            \renewcommand{\arraystretch}{1.05}
            \begin{tabular}{l|l|l}
                \toprule
                \textbf{Modulation} & $\boldsymbol{\mathcal{P}}$ & \textbf{Parameters} \\
                \midrule
                Arc & $\mathcal{P}_{\text{Arc}}$ & $p_{\text{Arc}}$: intensity \\
                Anticipation & $\mathcal{P}_{\text{Ant}}$ & $p_{\text{Ant}}$: intensity, $T_{\text{Ant}}$: time window  \\
                Slow In Slow Out & $\mathcal{P}_{\text{Slow}}$ & $\phi_{\text{Slow}}$: phase function \\
                Timing & $\mathcal{P}_{\text{Time}}$ & $p_{\text{Time}}$: intensity, $\phi_{\text{Time}}$: phase function \\
                Exaggeration & $\mathcal{P}_{\text{Exa}}$ & $p_{\text{Exa}}$: intensity \\
                Secondary Action & $\mathcal{P}_{\text{Sec}}$   &$p_{\text{Sec}}$: intensity \\
                Follow Through & $\mathcal{P}_{\text{Follow}}$   &$p_{\text{Follow}}$: intensity \\
                Randomization & $\mathcal{P}_{\text{Rand}}$   &$p_{\text{Rand}}$: intensity \\
                Interactive Goal & $\mathcal{P}_\text{Goal}$ & $g$: goal position \\
                \bottomrule
            \end{tabular}
            \caption{Modulation parameters}
            \label{tab:parameters}
            \vspace{-8mm}
        \end{table}

\section{Experiments}
The experiments include proof-of-concept demonstrations showcasing the modulation capabilities of individual animation principles, as well as demonstrations that combine multiple principles across various robot platforms with different kinematic configurations. These experiments highlight how diverse and complex expressions can be generated using a small set of meaningful parameters. Additionally, we conduct a user study to evaluate the effectiveness of our approach.

    \subsection{Proof-of-Concept Experiments}
\begin{figure}[t]
                \vspace{3mm}
                \centering
                \subfigure[$\mathcal{P}_{Arc}$]{
                \centering
                    \includegraphics[width=0.46\linewidth]{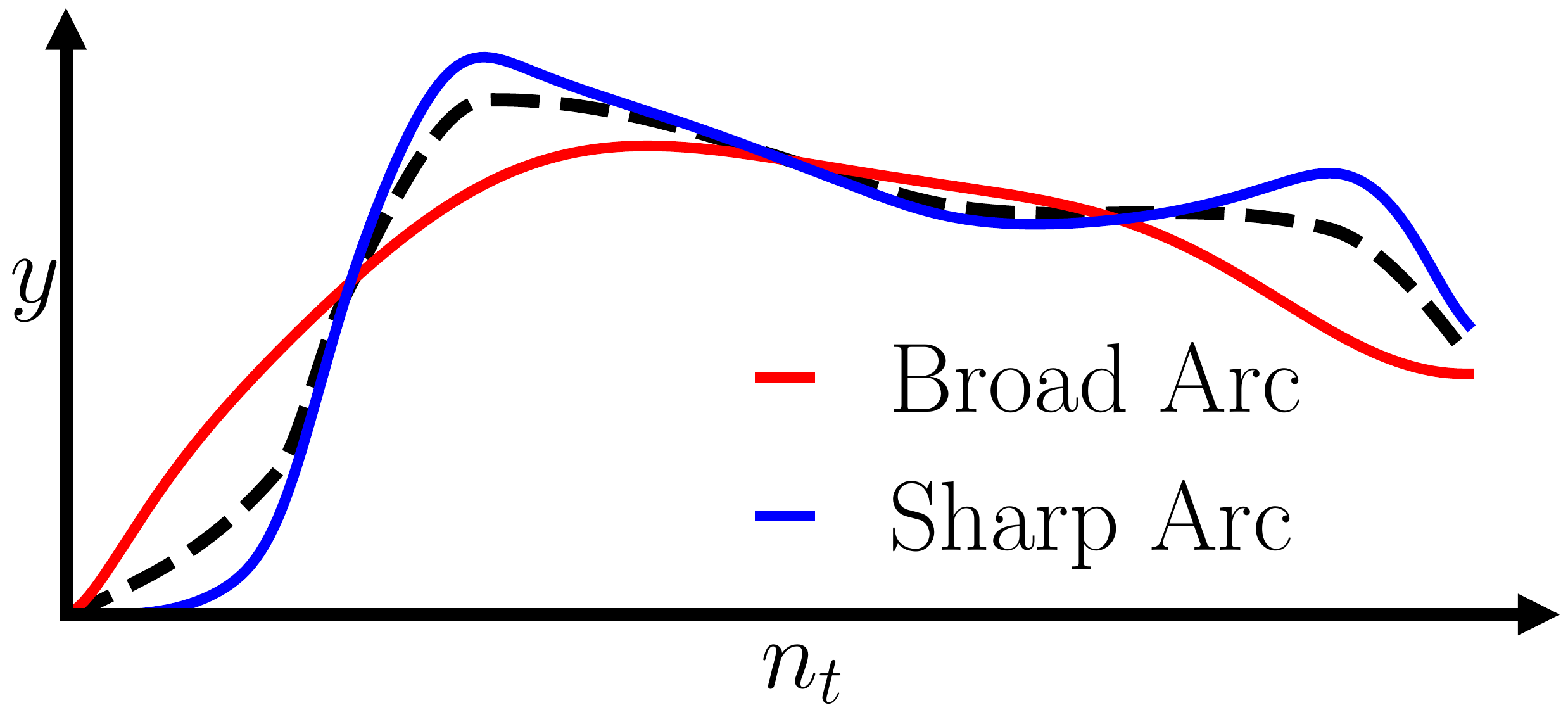}
                    \label{fig:arc}
                }                     
                \subfigure[$\mathcal{P}_{Ant}$]{
                \centering
                    \includegraphics[width=0.46\linewidth]{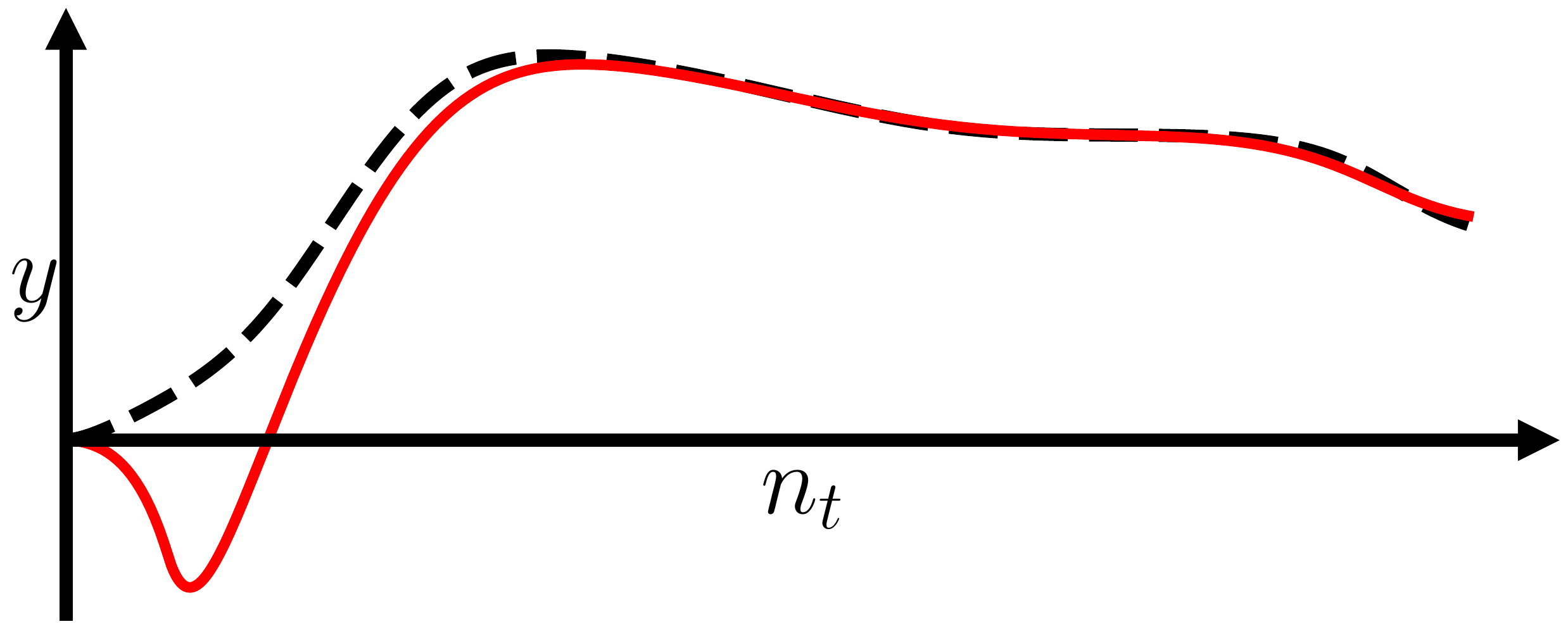}
                    \label{fig:anticipation}
                }
                \subfigure[$\mathcal{P}_{Time}$ Time Scaling]{
                \centering
                    \includegraphics[width=0.46\linewidth, trim={ 0 3mm 0 3mm},clip]{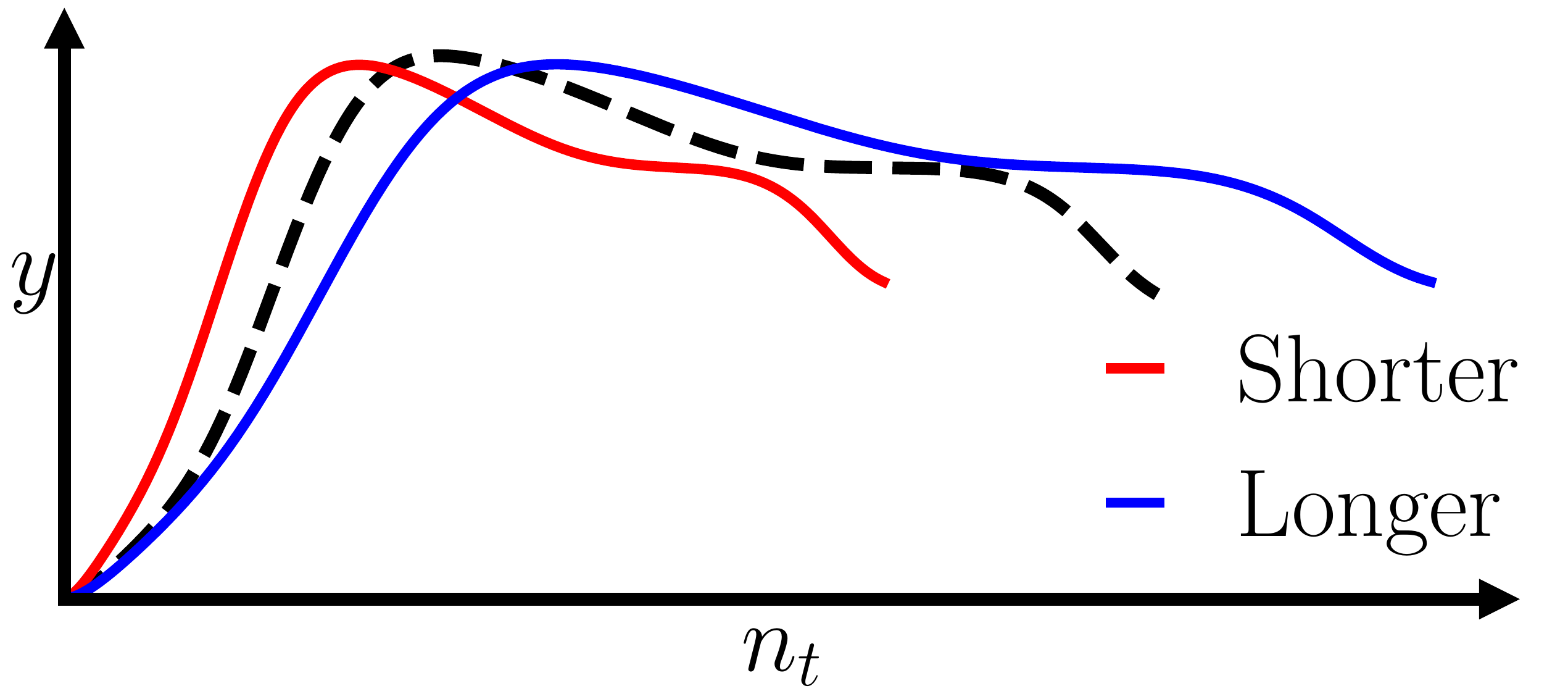}
                    \label{fig:timing_scaling}
                }%
                \subfigure[$\mathcal{P}_{Time}$ Progression, $\mathcal{P}_{Slow}$]{
                \centering
                    \includegraphics[width=0.46\linewidth, trim={ 0 3mm 0 3mm},clip]{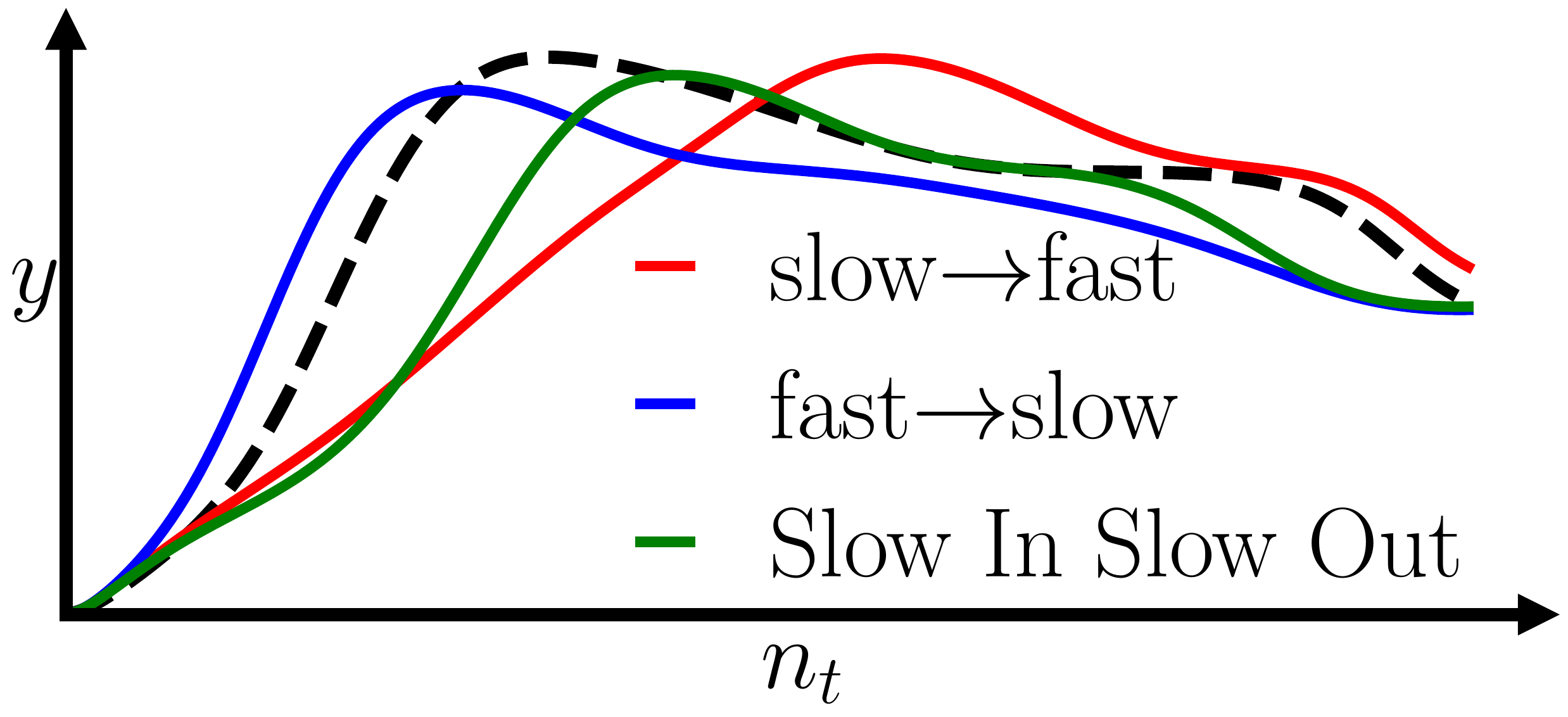}
                    \label{fig:timing_execution}
                }
                \subfigure[$\mathcal{P}_{Exa}$]{
                \centering
                    \includegraphics[width=0.46\linewidth, trim={ 0 3mm 0 3mm},clip]{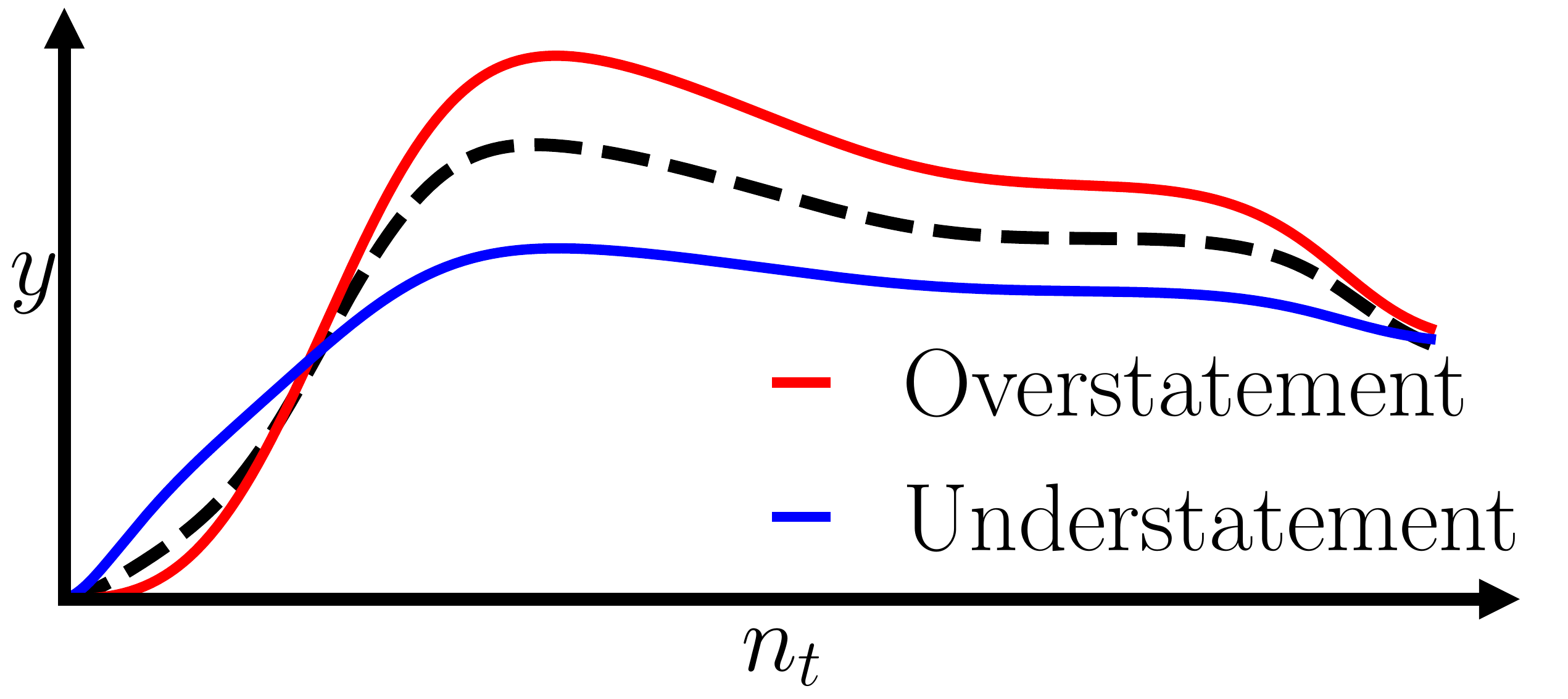}
                    \label{fig:exaggeration}                
                }
                \subfigure[$\mathcal{P}_{Sec}$]{
                \centering
                    \includegraphics[width=0.46\linewidth, trim={ 0 3mm 0 3mm},clip]{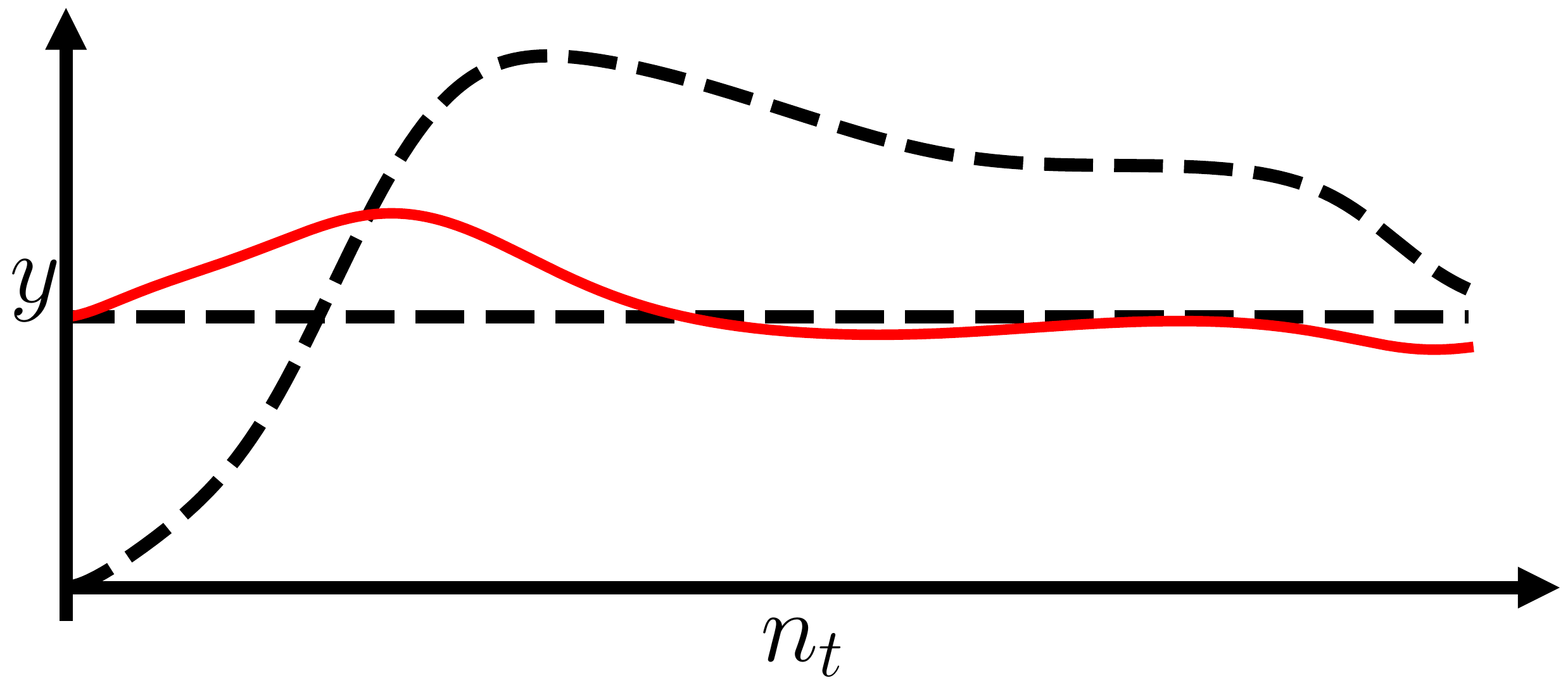}
                    \label{fig:secondary_action}
                }
                \subfigure[$\mathcal{P}_{Follow}$]{
                \centering
                    \includegraphics[width=0.46\linewidth, trim={ 0 3mm 0 3mm},clip]{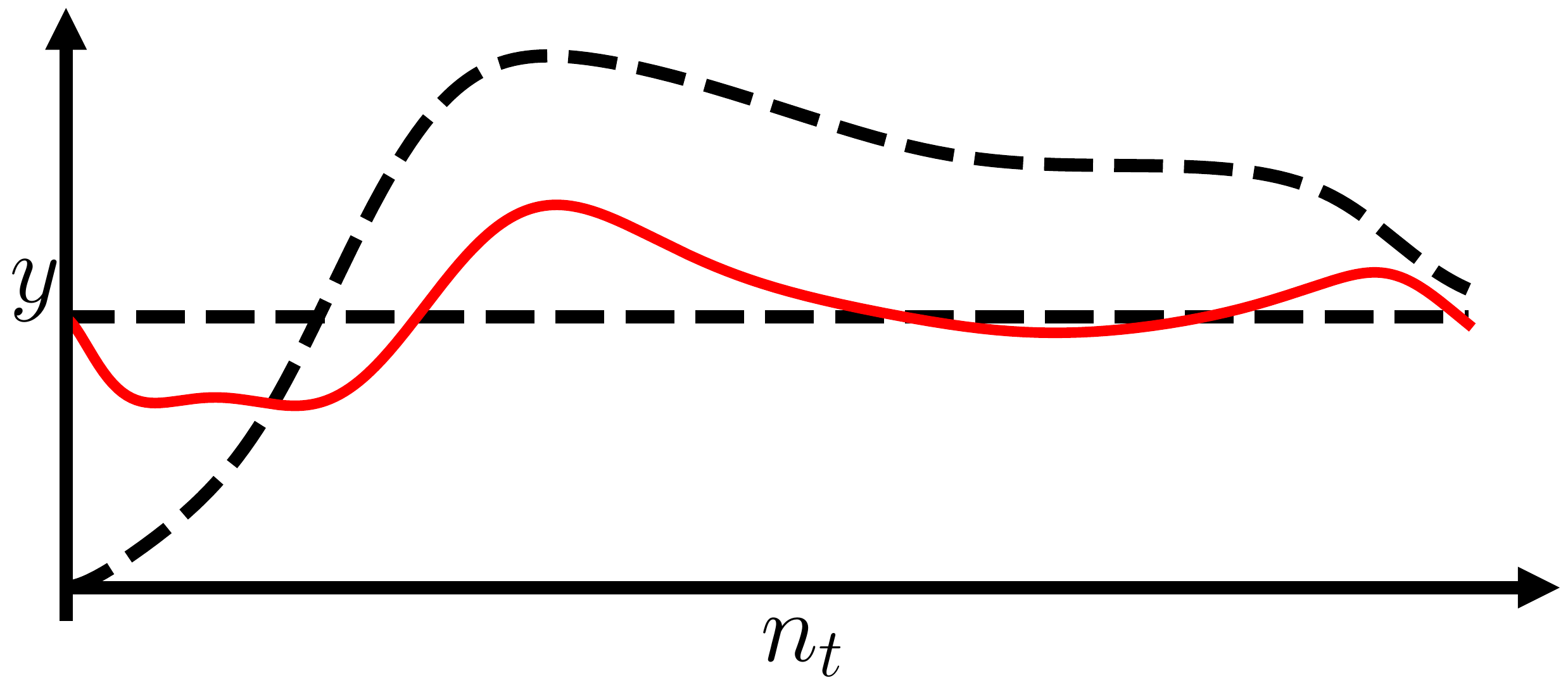}
                    \label{fig:follow_through}
                }
                \subfigure[$\mathcal{P}_{Rand}$]{
                \centering
                    \includegraphics[width=0.46\linewidth, trim={ 0 3mm 0 3mm},clip]{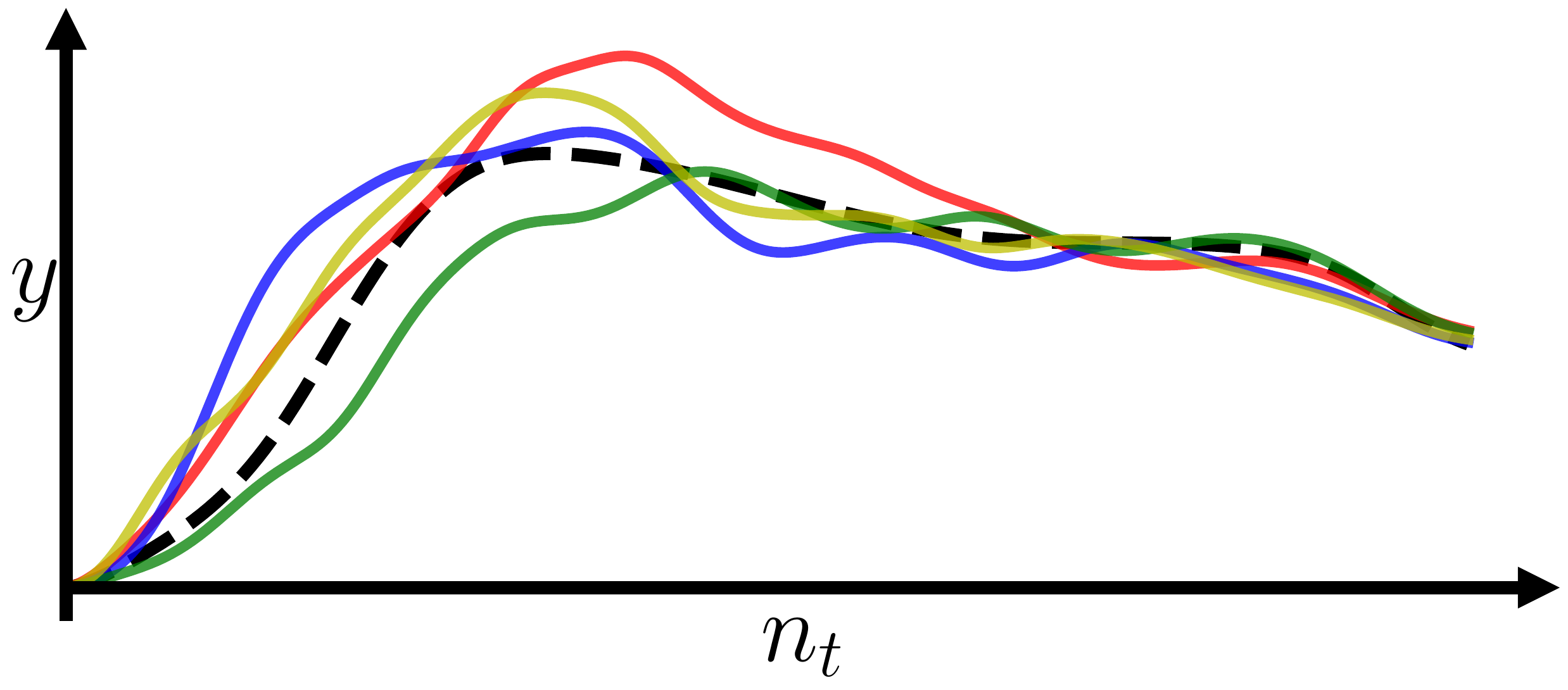}
                    \label{fig:slow_in_slow_out}
                }
                \caption{Modulation with individual principles. The demonstrated trajectories are shown as a dashed line.}
                \label{fig:1D}
                \vspace{-3mm}
            \end{figure}

        \emph{Single Dimension:}
            Fig.~\ref{fig:1D} shows modulations for each animation principle using a one-dimensional demonstration.
            The demonstration was chosen to cover relevant attributes: position value changes at the beginning and end of the trajectory, positive and negative velocities and accelerations, and steep or gentle slopes as key action features.

            Fig.~\ref{fig:1D}.a shows how different parameter values of $\mathcal{P}_{\text{Arc}}$ creates trajectories with a broader arc ($p_{\text{Arc}}=5$) and a sharper arc ($p_{\text{Arc}}=-5$).
            
            Fig.~\ref{fig:1D}.b: The anticipatory action induced by applying $\mathcal{P}_{\text{Ant}}$ (with $p_{\text{Ant}}=0.4$) directs the trajectory start into the opposite direction of the key action, pronouncing the motion.

            Fig.~\ref{fig:1D}.c: Scaling the execution time with $\mathcal{P}_{\text{Time}}$ for values $p_{\text{Time}}= \{ 0.75, 1.25 \}$ leads to an overall faster and slower motion behavior, where no key action features are lost.
            
            Using a phase modulation function $\phi_{\text{Time}}$ or $\phi_{\text{Slow}}$, progress modulation can be achieved. Fig.~\ref{fig:1D}.d shows progression modulation that causes the motion to be slower for ``slow'' sectors and faster for ``fast'' sectors for a simple sector-based choice for $\phi_{\text{Time}}$. It also shows motion for $\mathcal{P}_{\text{Slow}}$ for slow progress at the beginning and the end with an inverted sigmoid as a phase modulation function.  
            
            Fig.~\ref{fig:1D}.e: Modulation for  $\mathcal{P}_{\text{Exa}}$ leads to overstatement ($p_{\text{Exa}}=1.5$) or understatement ($p_{\text{Exa}}=0.5$) of the motion's key action features in a way that they get either amplified or attenuated. For $p_{\text{Exa}}=0$, the forcing term disappears, and the motion becomes a feature-less movement to the goal $g$.
            
            Fig.~\ref{fig:1D}.f: The modulated trajectory of $\mathcal{P}_{\text{Sec}}$ (with $p_{\text{Sec}}=0.05$) clearly shows how the values of the \textit{target} dimension are guided to follow those of the \textit{source} dimension, further amplifying the key action features.
            Similarly, for $\mathcal{P}_{\text{Follow}}$ in Fig.~\ref{fig:1D}.g, the course of the \textit{target} dimension with $p_{\text{Follow}}=3.0$ shows a lagging behavior with swinging over the turning point following the primary action. 
            
            Finally, Fig.~\ref{fig:1D}.h shows multiple randomized trajectories for $p_{\text{Rand}}=0.5$.

\begin{figure}[t]
            \vspace{4mm}
                \centering
                \hspace{-15pt}
                \subfigure[3D $\mathcal{P}_{\text{Arc}}$]{
                    \includegraphics[width=0.375\linewidth, trim={ 0 15mm 0 15mm},clip]{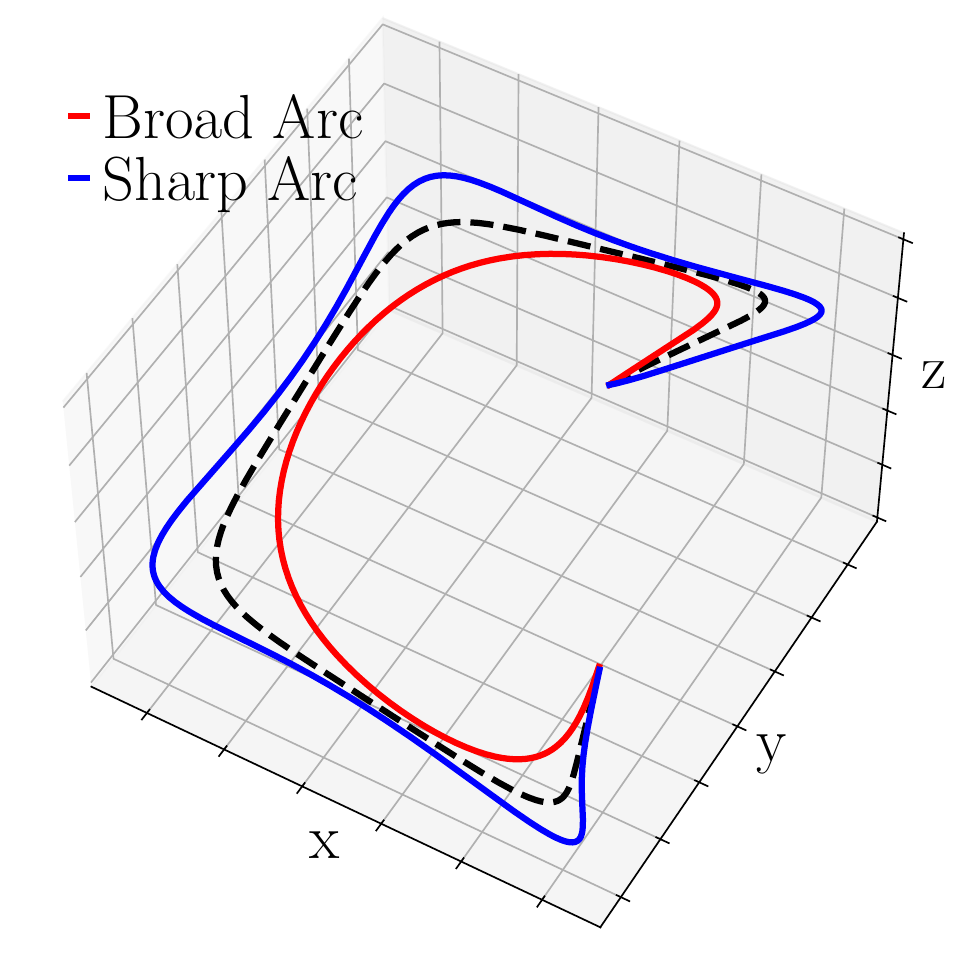}
                    \label{fig:arc_3D}
                }\hspace{-15pt}
                \subfigure[3D $\mathcal{P}_{\text{Exa}}$ and $\mathcal{P}_{\text{Ant}}$]{
                    \includegraphics[width=0.375\linewidth, trim={ 0 15mm 0 15mm},clip]{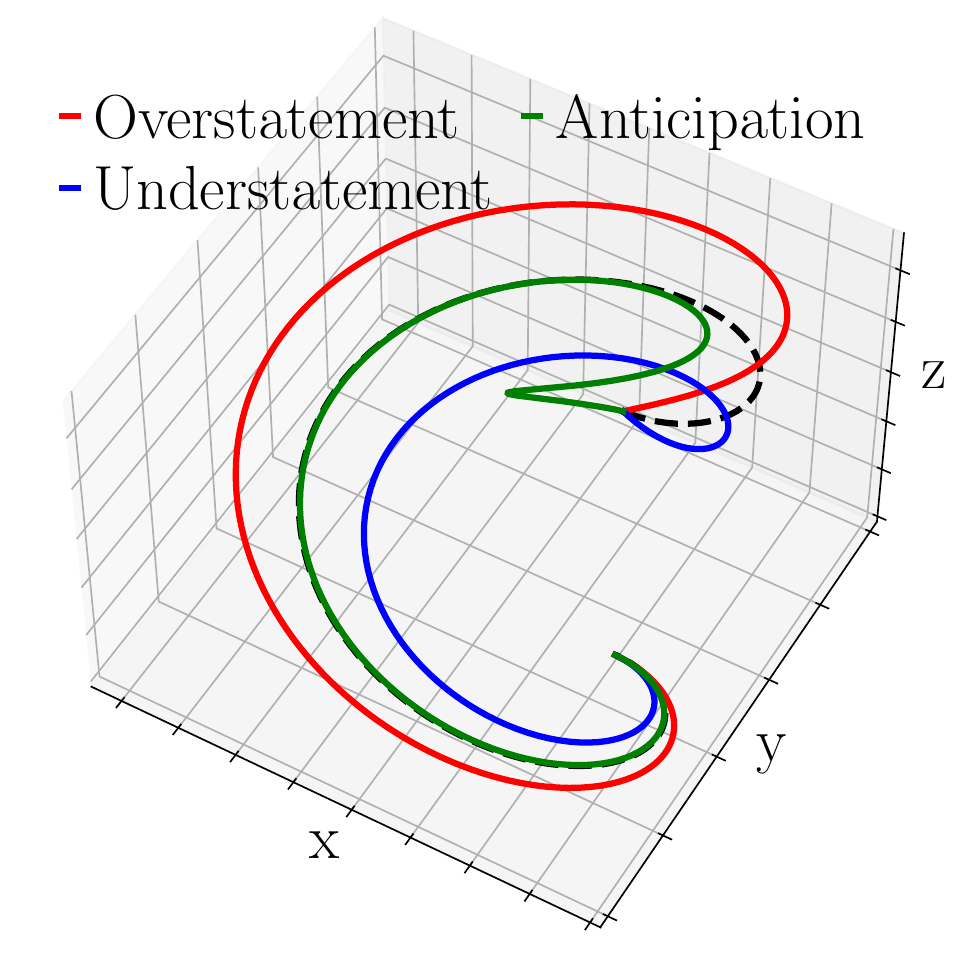}
                    \label{fig:E_A_SA_FT_3D}
                }\hspace{-10pt}
                \subfigure[\textit{Pepper} comb. $\mathcal{P}$]{
                    \includegraphics[width=0.26\linewidth, trim={ 0 0 0 0},clip]{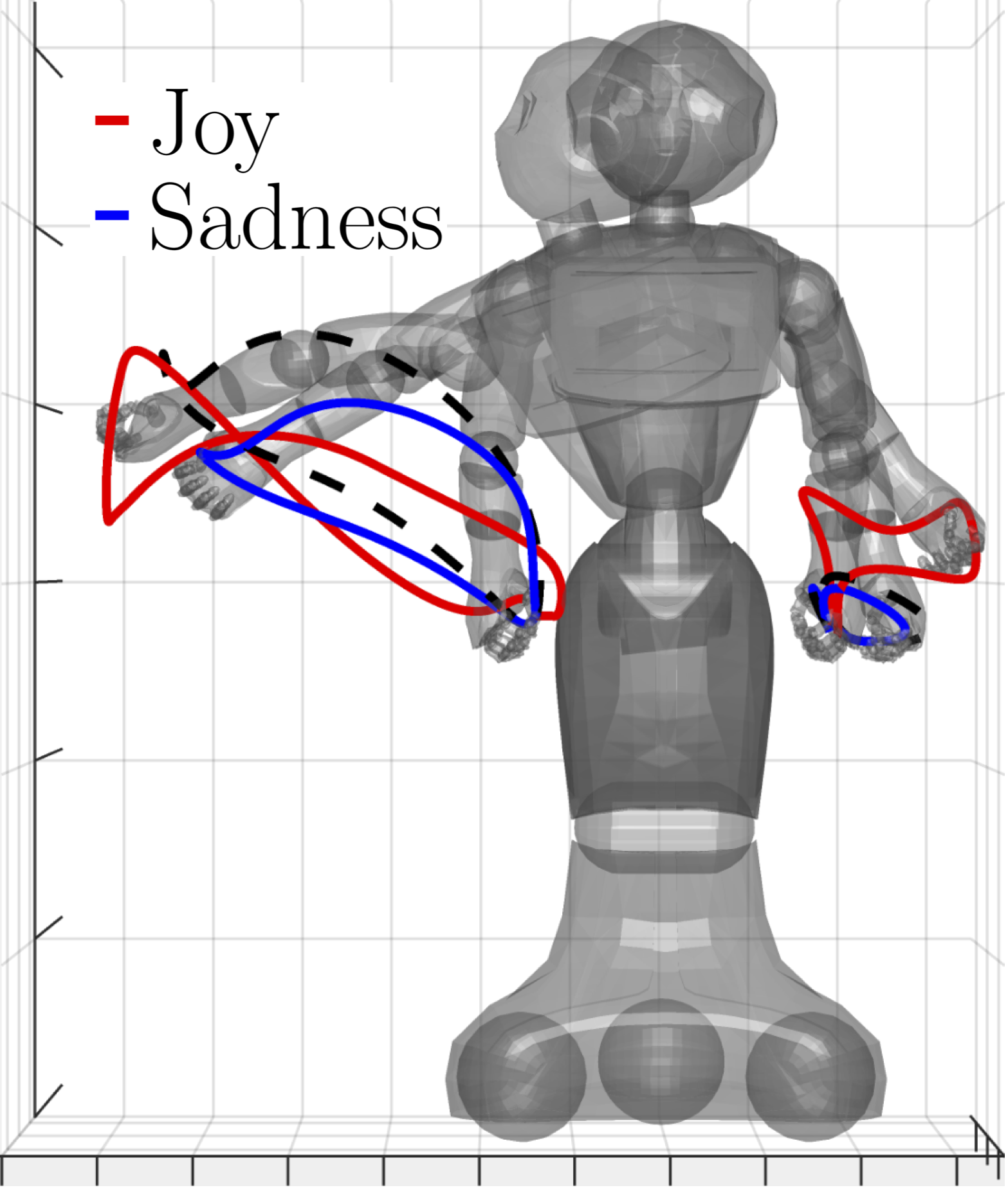}
                    \label{fig:updated_pepepr}
                }
                \hspace{-15pt}
                \caption{Selected modulations in higher dimensions. The demonstrated trajectory is shown as a dashed line.}
                \label{fig:3D_plots}
                \vspace{-3mm}
            \end{figure}

        \emph{Higher Dimensions:}
            In Fig.~\ref{fig:3D_plots}.a and Fig.~\ref{fig:3D_plots}.b, we show modulations of selected animation principles when increasing the system's dimensionality to 3D Cartesian space. %
            As a demonstration, we use two trajectories that sample the surface of a sphere with linear interpolation from top to bottom, roughly in Fig.~\ref{fig:3D_plots}.a and densely in Fig.~\ref{fig:3D_plots}.b.
            For $\mathcal{P}_{\text{Arc}}$, different values for $p_{\text{Arc}}$ modulate the arc of the resulting trajectory to be broader or sharper. Also, for the other principles, the modulation characteristics retain the properties shown in one dimension, with $\mathcal{P}_{\text{Ant}}$ showing an anticipatory countermovement in the beginning.
            
    \subsection{Robot Demonstrations}
\begin{figure}[t]
            \vspace{3mm}
            \centering            
            \includegraphics[height=5cm]{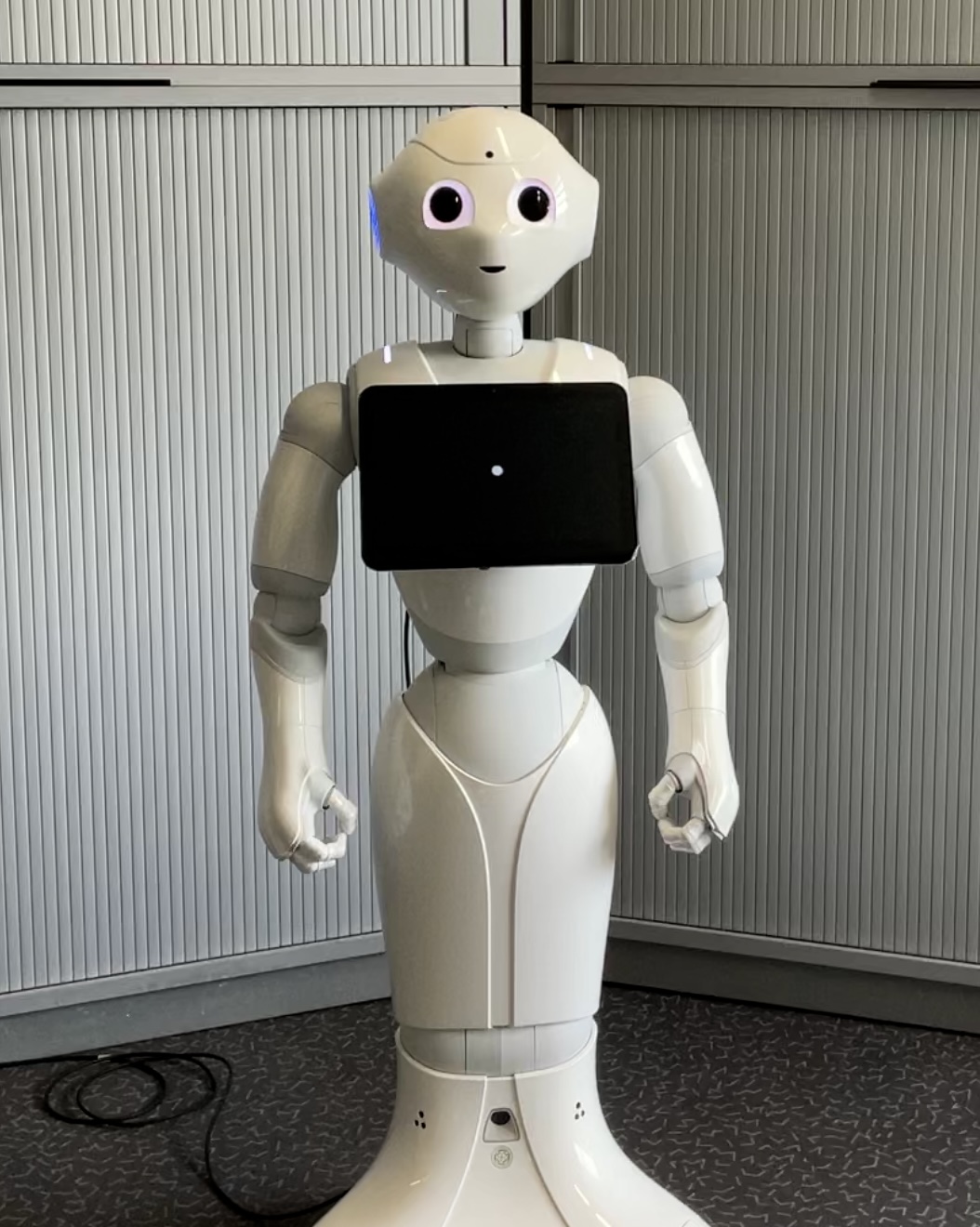}
            \includegraphics[height=5cm]{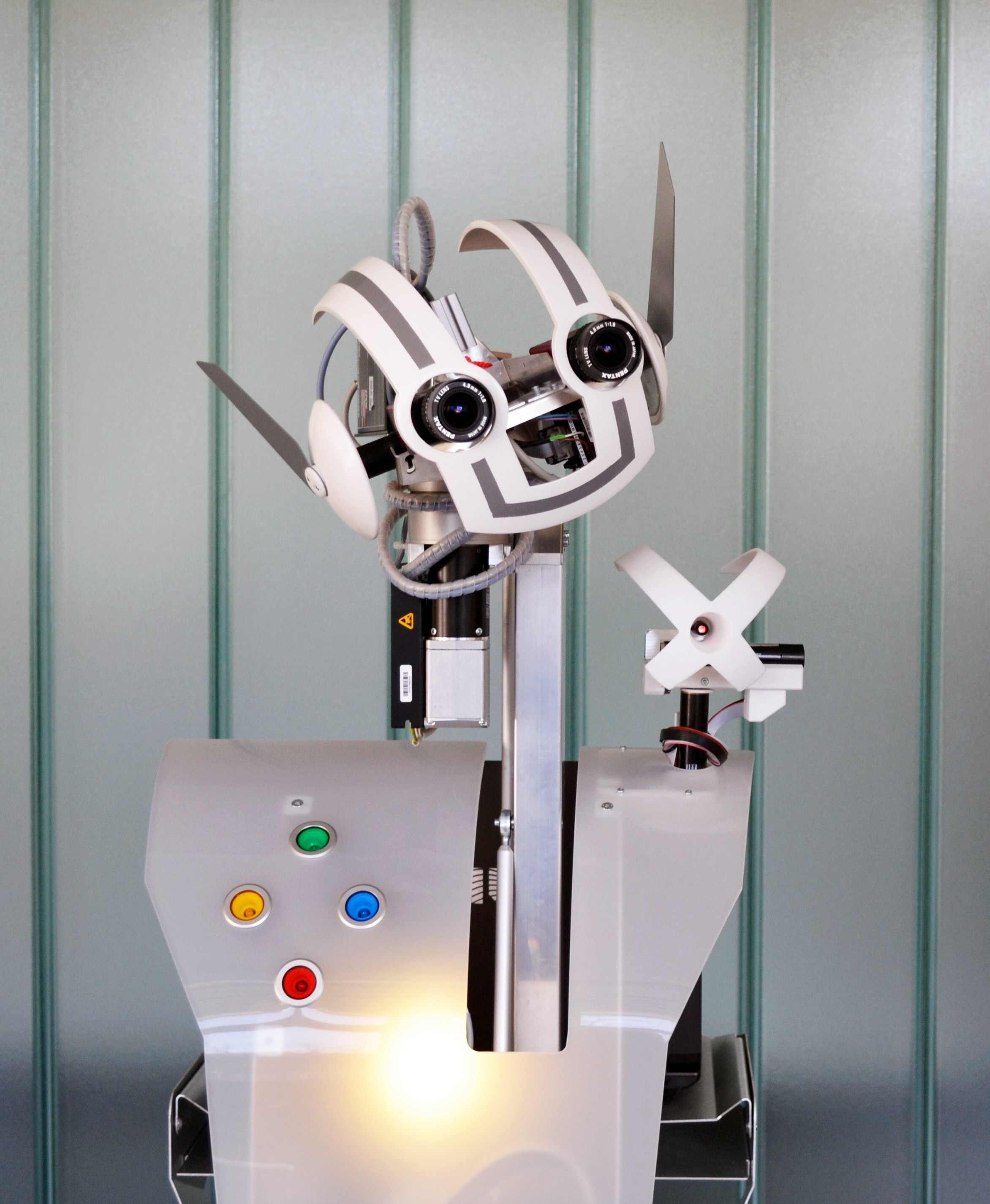}
           \caption{Two of the robots used in our experiments: \textit{Pepper} (left), \textit{Daryl} (right)}
            \label{fig:real_robots}
            \vspace{-3mm}
        \end{figure}

        We implemented our approach on three different robots, in simulation on a \textit{LBR iiwa} by KUKA, and on two robot platforms, \textit{Pepper} and \textit{Daryl}, see Fig.~\ref{fig:real_robots} and the supplementary video.
        
        First, we demonstrate our approach by modulating simple, expressive motions with different animation principles.
        
        \textit{Daryl} is a mildly humanized interactive mobile robot platform with ten DoFs. We only use its head's tilt joint to execute a nodding motion. Nodding is a fairly universal way to indicate agreement or non-verbal acceptance and conveys meaningful cues when speed or extent are varied. The motion is modulated in three ways: with $\mathcal{P}_{\text{Time}}$ we make the motion faster, with $\mathcal{P}_{\text{Exa}}$ we exaggerate the motion, and with $\mathcal{P}_{\text{Rand}}$, we induce the robot's neutral configuration with small random movements for a more lifelike appearance.

        Next, we use \textit{Pepper} and a simple arm-raising gesture to demonstrate $\mathcal{P}_{\text{Ant}}$. The anticipatory motion amounts to an inverted shoulder joint rotation at the beginning of the execution, which leads to a retraction of the arm behind the body before it is lifted to complete the raising of the arm.
        We also show our DMP approach's inherent goal-orientedness with online head orientation adaptions during a nodding motion.

        Further, the \textit{LBR iiwa}, a non-anthropomorphic robot arm with seven DoFs, the model is trained with a jagged demonstrated trajectory and modulated with $\mathcal{P}_{\text{Arc}}$ to make the motion look more natural, smooth and with gradual direction changes.   

        We demonstrate that our approach applies to whole-body motion on complex robot kinematics. We use \textit{Pepper} with its 17 DoFs corresponding to the joints responsible for the movements of the torso, the head, and the arms.
        Gestures taken from \textit{Pepper}'s library of handcrafted animations \cite{softbank_robotics_europe_pepper_2019} and learned using our DMP approach are modulated with individual principles and combinations of multiple principles. This is described in more detail in the next section, which covers the user study.              

        Motion in all of the above experiments aligns with the definitions of the respective principles and appears consistent with the qualitative expectations based on human intuition.

\subsection{User Study}

We further designed a controlled laboratory study to evaluate if and how well participants could perceive the different modulations. 
The study consisted of two parts: The first part studied participants' ability to recognize the emotional expression of a modulated gesture correctly; the second part studied whether different execution intensities resulted in the corresponding desired subjective intensity ratings.
We recruited 34 participants (10 female) aged between 20 and 60 years using university mailing lists and notice boards.
All participants completed both parts of the study.

\paragraph{Emotional Expression} In the first part of the study, we used our method to generate different modulations of an interaction gesture of the \textit{Pepper} robot to express different emotional expressions:
Specifically, we manually created six modulations that carried the following emotional expressions: the four primary emotions Joy, Sadness, Anger, and Fear, and two secondary emotions Shame and Hurry.
Fig.~\ref{fig:3D_plots}.c shows a sample subset of emotional expressions for a part of the gesture.
Using these modulations, we studied hypothesis
\textbf{H1:} {\em The extent to which participants recognize the emotional expression of a modulated gesture is above chance.}
Users were shown videos of the six variations and asked to answer to which extent they perceived each of the six possible options on a 5-point linear scale where the lowest perception score was ''not at all``, and the highest was ''very strongly``.\\
We compared the aggregated data for the target emotional expression with the chance level using a one-sample t-test.

We found that in all instances, participants could recognize the intended emotional expression above chance with statistical significance ($p<.05$). 
The data also shows that motion with a certain intended emotional expression can be perceived as multiple emotional expressions. This aligns with the findings in \cite{thomas_illusion_1995} and \cite{wallbott1998bodily}, which suggest that certain emotional expressions share basic motion attributes.

\paragraph{Expression Intensity} For the second part of the study, we generated expressive modulations for individual motions by scaling their intensities.
Using these high- and low-intensity motions, we studied two hypotheses,
\textbf{H2.1:} {\em The intensity ratings of participants seeing high-intensity motion are higher than those seeing low-intensity motion,} and 
\textbf{H2.2:} {\em The intensity ratings of participants seeing low-intensity motion are lower than those seeing high-intensity motion.}
Participants were first shown medium-intensity motion and then the videos for intended low- or high-intensity motion.
For all variations, participants were asked to choose ''lower intensity`` or ''higher intensity`` based on their perceived intensity of each variation.

TABLE~\ref{tab:study_part_2} shows normalized confusion matrices for individual principles comparing modulation intensities $I_M$ versus perceived intensities $I_P$.
The data was analyzed using one-sided independent t-tests. As can be seen from the table, the collected data supports both hypotheses with statistical significance ($p<.05$). This confirms that the scales of the parameters for individual principles follow the intended intuition, which is particularly important as the parameters serve as the main tuning factor for using our method.

        \begin{table}[]
            \centering
            \begin{tabular}{c|cc|cc|cc|cc|cc|cc}
                \toprule
                &
                \multicolumn{2}{c|}{$\mathcal{P}_{\text{Arc}}$} &
                \multicolumn{2}{c|}{$\mathcal{P}_{\text{Ant}}$} &
                \multicolumn{2}{c|}{$\mathcal{P}_{\text{Time}}$} &
                \multicolumn{2}{c|}{$\mathcal{P}_{\text{Exa}}$} &
                \multicolumn{2}{c|}{$\mathcal{P}_{\text{Sec}}$} &
                \multicolumn{2}{c}{\makebox[0pt]{$\mathcal{P}_{\text{Follow}}$}} \\
                \midrule
                \diagbox[width=3em, height=2em]{\makebox[0pt]{$I_M$}}{\makebox[0pt]{$I_P$}} &
                \multicolumn{1}{c}{$\uparrow$} & $\downarrow$ &
                \multicolumn{1}{c}{$\uparrow$} & $\downarrow$ &
                \multicolumn{1}{c}{$\uparrow$} & $\downarrow$ &
                \multicolumn{1}{c}{$\uparrow$} & $\downarrow$ &
                \multicolumn{1}{c}{$\uparrow$} & $\downarrow$ &
                \multicolumn{1}{c}{$\uparrow$} & $\downarrow$ \\
                \midrule
                high &
                \makebox[4pt]{$\scriptstyle\mathbf{0.94}$} & \makebox[4pt]{$\scriptstyle0.06$} &
                \makebox[4pt]{$\scriptstyle\mathbf{0.79}$} & \makebox[4pt]{$\scriptstyle0.21$} &
                \makebox[4pt]{$\scriptstyle\mathbf{1.00}$} & \makebox[4pt]{$\scriptstyle0.00$} &
                \makebox[4pt]{$\scriptstyle\mathbf{1.00}$} & \makebox[4pt]{$\scriptstyle0.00$} &
                \makebox[4pt]{$\scriptstyle\mathbf{0.94}$} & \makebox[4pt]{$\scriptstyle0.06$} &
                \makebox[4pt]{$\scriptstyle\mathbf{0.56}$} & \makebox[4pt]{$\scriptstyle0.44$} \\
                low &
                \makebox[4pt]{$\scriptstyle1.00$} & \makebox[4pt]{$\scriptstyle\mathbf{1.00}$} &
                \makebox[4pt]{$\scriptstyle0.26$} & \makebox[4pt]{$\scriptstyle\mathbf{0.74}$} &
                \makebox[4pt]{$\scriptstyle0.00$} & \makebox[4pt]{$\scriptstyle\mathbf{1.00}$} &
                \makebox[4pt]{$\scriptstyle0.00$} & \makebox[4pt]{$\scriptstyle\mathbf{1.00}$} &
                \makebox[4pt]{$\scriptstyle0.03$} & \makebox[4pt]{$\scriptstyle\mathbf{0.97}$} &
                \makebox[4pt]{$\scriptstyle0.29$} & \makebox[4pt]{$\scriptstyle\mathbf{0.71}$} \\
                \bottomrule
            \end{tabular}
            \caption{Normalized confusion matrices for intensity-scaled principles. Arrows denote higher and lower perceived intensities based on high and low modulation intensities.}
            \label{tab:study_part_2}
            \vspace{-5mm}
        \end{table}

        \begin{figure}
            \vspace{4mm}
            \centering            
            \includegraphics[width=0.95\linewidth, trim={ 0 0 0 6pt},,clip]{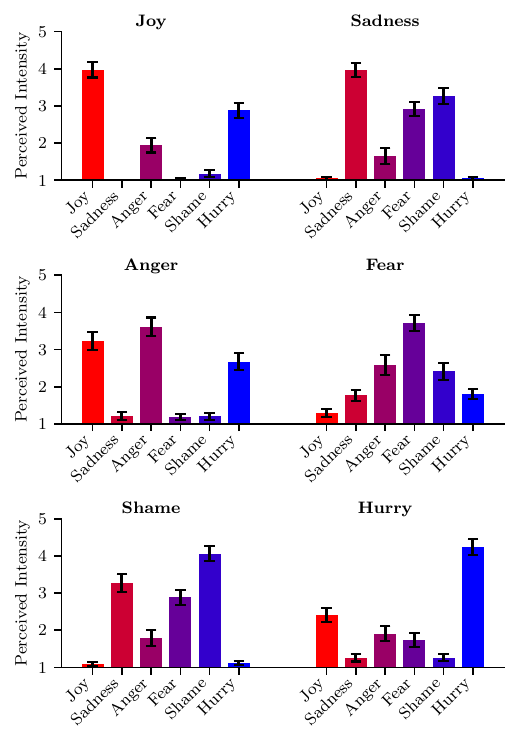}
            \vspace{-5mm}
            \caption{Average perceived intensities for all emotional expressions. Error bars indicate standard error.}
            \label{fig:study_part_1}
        \end{figure}

\section{Conclusion}

    In this work, we present a method for generating expressive robot motion by leveraging Dynamic Movement Primitives to systematically compose and modulate reusable motion skills along the meaningful dimensions of the Principles of Animation. Our approach is data-efficient, learnable, explainable, modulable, online adaptable, and composable.
    Experiments in simulation, real robot platforms, and a user study demonstrate that our method provides an intuitive and effective tool for modulating complex whole-body motion across different kinematic configurations, using both individual principles and combinations of multiple principles.
    Future work will focus on learning modulation parameters from data, as well as imitation learning of human motion from visual cues of a body pose tracker and the application of the approach for HRI tasks.

\balance
\bibliographystyle{IEEEtran}
\bibliography{DMP_references}

\end{document}